\documentclass{article}
\usepackage[utf8]{inputenc}

\usepackage[margin=.8in]{geometry}
\newtheorem{theorem}{Theorem}
\usepackage{parskip}
\usepackage{newpxtext}
\usepackage{newpxmath}
\usepackage{xparse}
\usepackage{amsmath}
\usepackage{mathtools}
\usepackage{microtype}
\usepackage{enumitem}
\usepackage{xcolor}
\usepackage{adjustbox}
\usepackage{varwidth}
\usepackage{fvextra}
\usepackage{hyperref}
\usepackage{graphicx}
\usepackage{placeins}
\usepackage{color}

\usepackage[
backend=biber,
style=alphabetic,
sorting=ynt
]{biblatex}

\DeclareMathOperator{\argmin}{argmin}

\DeclareMathOperator{\MIP}{MIP}

\title{A Circuit Complexity Formulation of Algorithmic Information Theory}
\date{}

\begin{document}
\author{Cole Wyeth, Carl Sturtivant}
\maketitle

\begin{abstract}
    Inspired by Solomonoff's theory of inductive inference \cite{SOLOMONOFF19641}, we propose a prior based on circuit complexity. There are several advantages to this approach. First, it relies on a complexity measure that does not depend on the choice of UTM. There is one universal definition for Boolean circuits involving an universal operation (e.g. NAND) with simple conversions to alternative definitions (with AND, OR, and NOT). Second, there is no analogue of the halting problem. The output value of a circuit can be calculated recursively by computer in time proportional to the number of gates, while a short program may run for a very long time. Our prior assumes that a Boolean function (or equivalently, Boolean string of fixed length) is generated by some Bayesian mixture of circuits. This model is appropriate for learning Boolean functions from partial information, a problem often encountered within machine learning as "binary classification." We argue that an inductive bias towards simple explanations as measured by circuit complexity is appropriate for this problem.
\end{abstract}

\section{Introduction}

Solomonoff's general theory of inductive inference \cite{SOLOMONOFF19641} aims to solve the problem of sequence prediction by considering a Bayesian mixture of all computable explanations for the subsequence observed so far. This naturally give rise to the notion of Kolmogorov complexity of a string x, the shortest computer program producing output x. Inspired by this construction, we propose a prior based on a Bayesian mixture of circuits, with the circuit complexity taking the place of the Kolmogorov complexity. This overcomes some of the limitations of the Kolmogorov complexity, such as its dependence on the choice of Universal Turing Machine (UTM). Our model is "parameterized" by our choice of universal logic gate(s), but none of our main results depend on this choice significantly. Our approach also circumvents the halting problem, since the output of a small circuit can always be calculated quickly. Solomonoff's theory is discussed further in section \ref{SolomonoffInductionDefinitions}.

Our prior is appropriate for learning Boolean functions from partial information. Within the machine learning community, this is referred to as supervised learning for binary classification. The inputs can be viewed as features, and we are provided with some input/output pairs (the training data). Our task is to predict the output on unseen inputs (the test data). Since this is clearly an underspecified problem, some inductive bias is necessary; our prior prefers explanations (Boolean functions) with low circuit complexity (as well as those computed by many circuits). The circuit prior is constructed explicitly in section \ref{Defs}. The symbols introduced in this section appear for convenience in the appendix, section $\ref{Glossary}$.

In section \ref{MIP}, we propose a prediction scheme based on the smallest circuit consistent with the training data. This is a natural approximation to the circuit prior, ignoring multiplicity of explanations and predicting based on the best one. The problem is also reduced from an infinite sum to a search. In section \ref{MIP Proof} we derive upper and lower bounds on the number of errors we should expect to make with this approximation. These results are loosely analogous to the coding theorem in algorithmic information theory (AIT) or to a deterministic case of the minimum description length (MDL) principle. 

Finally, we suggest some directions for future work, including investigations of infinite sequence prediction.

\section{Background}

\subsection{Learning circuits}

The problem of learning circuits (and other models of computation) from examples in polynomial time is studied in the field of learning theory, as initiated by Valiant \cite{learnability}. His model is well suited for studying the feasibility of finding a circuit that agrees with an unknown Boolean function from positive and negative examples (that is, inputs for which the function is 1 or 0 respectively) with high probability. This is intuitively the "frequentist dual" of our approach. Instead of considering a fixed Boolean function, which can be represented by a circuit, and judging the performance of a learning algorithm on test data as a random variable depending on randomly selected training data, we study Bayesian optimal prediction based on a latent circuit with some prior distribution. This approach is more closely in line with the usual assumptions of AIT. 

Our approach is actually closer in spirit to structure learning in the study of Bayesian networks \cite{BayesianNetworkStructure}. A circuit can be viewed as a special (deterministic) type of Bayesian network, since both are DAGs (though we only have access to values for the input and output vertices). Often a maximum a posteriori (MAP) estimate of network structure is found with a greedy search method \cite{CausalDiscovery}, instead of using the full Bayesian mixture. The MDL predictor we investigate in section \ref{MIP}, MIP, can be thought of as a MAP estimate. 

\subsection{Solomonoff's theory of prediction} \label{SolomonoffInductionDefinitions}

The Kolmogorov complexity of a string x, denoted K(x), is defined as the shortest pair <M,u> encoding a TM M and an input u such that M run on u halts with x on the tape. K(x) is defined with reference to a prefix machine with self-delimiting programs, and this definition is extended to the conditional complexity K(x|y) by providing y as an additional input. Since we are interested in computing strings by circuit, and a circuit produces a bit of a string given its index as input, it is perhaps more pertinent to consider programs which perform the same task. In this direction, Eric Allender has introduced a version of time complexity called KT(x), which is the smallest sum of length and runtime for a program that produces a bit of x given its index \cite{WorldsCollide}. 

Solomonoff constructed a prior by sampling programs (on a fixed UTM) instead of circuits. In particular, a UTM U is run with its input tape initialized with an i.i.d. infinite sequence of 0's and 1's, obtained by "coin flips" (a Bernoulli process with  $\frac{1}{2}$ probability of success). This sequence is interpreted as a program for U. A distribution over $\mathbb{B}^n$ can then be obtained from the output of this UTM:
\[
P_d(x) = \sum_{U(p) = x} 2^{-l(p)}
\]
This is the discrete version of Solomonoff's universal a priori distribution. More commonly considered is the continuous version, in which U is allowed to output bits forever without halting:
\[
P_c(x) = \sum_{U(p) = x*} 2^{-l(p)}
\]
This continuous version can be interpreted as a semi-measure on [0,1] by treating bit strings as the binary expansions of real numbers. The * here stands for any finite or infinite bitstring, contrary to the meaning in this paper. 

The discrete version is more analogous to our case. It is a theorem of algorithmic information theory that the (log inverse) probability of a string according to $P                                                                                                                                                    _d$ is within a constant of its Kolmogorov complexity, which is called the coding theorem \cite{livitanyi}. The heuristic of relying on the shortest consistent explanation is a good approximation to Solomonoff's induction scheme. 

Many variations of Solomonoff's theory have been studied. Particularly notable is Schmidhuber's "Speed Prior" \cite{SpeedPrior}, which is similar to our approach in that it penalizes long running computations. However, it is still based on TMs instead of circuits, which strongly suggests it retains some of the disadvantages of Solomonoff induction (including the lack of known algorithms for effectively searching the space of TMs).  

\section{Definitions} \label{Defs}

The naive extension of a TM computing a string to circuits would be to define the complexity of a string x as the "size" (or some other complexity measure involving number of inputs and/or gates) of the smallest circuit that outputs x on some input. No matter how this is formulated it fails immediately because 2 inputs are always sufficient: 0 and 1 can be wired directly to each 0 and 1 position in x. The bits 0 and 1 can even be hardwired given a single input. It is not sensible to produce the bits of x with the output gates of a circuit. 

In fact, the natural idea is to choose the smallest circuit that produces the bits of x given their index within x. Such a circuit has $\lceil \lg |x| \rceil$ input gates, along with 0 or more additional gates, and one designated output gate (which may also be an input). Of course, the length of x may not be a power of 2, so it's immediately clear that we will have some freedom in our choice of circuit for x, since the values for indices beyond it's length are ambiguous. With this in mind, we will map strings to partial functions from $\{ 0, 1 \}^n \rightarrow \{ 0, 1 \}$, as follows:

Let A be the alphabet $\{ 0, 1, * \}$ where * is interpreted as an undetermined bit. For any string $x \in A^n$, we will interpret x as a subset of $\mathbb{B}^n$ that includes all fillings of stars with 0 or 1. We will restrict our focus to n a power of 2 so that x can be interpreted as a partial function from $\lg n$ input bits to n outputs, and any $s \in x$ as a function. For now we will fix n, and a string of length less than n is implicitly padded with stars to length n. We will occasionally use the notation that $x_{i:j}$ represents the substring of x between the indices of i and j (inclusive), and that $x_{<k}, x_{\leq k}$ represent prefix of x up to index k, exclusive and inclusive respectively. 

Now, we will construct a measure on $\mathbb{B}^n$ based on a model of randomly sampling a circuit and computing a string from that circuit. Since we can think of this process as calculating each bit position of a string by passing its index as input to the circuit, we will sometimes refer to this as an indexing circuit for the string. 

Formally, let $\Lambda$ be the set of circuits. We may define these as directed acyclic graphs with a set of $\lg n$ designated inputs and a designated output, requiring fan-in 2 for non-input vertices. The output may also be an input. Each vertex is interpreted as a NAND gate. 

We will denote string computation by $\pi : \Lambda \rightarrow \mathbb{B}^n$, so that if a circuit $c \in \Lambda$ computes a string $s \in \mathbb{B}^n$ we write $\pi(c) = s$. Let $\lambda$ be a measure on $\Lambda$. Then, if we are attempting to learn a Boolean function, and our prior information tells us that it was computed by some circuit sampled from $\Lambda$ with probability given by $\lambda$, our prior on $\mathbb{B}^n$ is $\mu = \lambda \circ \pi^{-1}$, the pushforward $\pi_*(\lambda)$ of $\lambda$ to $\mathbb{B}^n$. We will call $\mu$ the circuit prior (or more accurately, a circuit prior derived from $\lambda$). Intuitively, it expresses a belief that strings are more likely if they are computed by many relatively likely circuits. This can be justified by the Principle of Multiple Explanations, dating back to Epicurus \cite{UniversalInduction}. The question remains of how to construct $\lambda$. Given two alternative explanations for the same data, Ockham's razor tells us to prefer the simpler one. In the context of circuits, simplicity corresponds to circuit size. With a fixed fan-in, this can be defined as the number of gates. Our convention is not to count the input gates since they appear in every circuit. We will let |c| denote the size of a circuit. Let $\mathcal{G}$ be the set of possible circuit sizes (an alias for $\mathbb{N}$). Then as a function, we have || : $\Lambda \rightarrow \mathcal{G}$. We can express our preference for simple circuits by writing $\lambda = f \circ ||$, for some $f : \mathcal{G} \rightarrow \mathbb{R}$. Then in full, $\mu : \mathbb{B}^n \rightarrow \mathbb{R} = f \circ || \circ \pi^{-1}$.

We will investigate effectiveness of this heuristic for $\mu$ - that is, how badly can we do by using the smallest circuit that explains the bits of a string x we have seen? This will form the first contribution we are aware to a theory of prediction based entirely on circuits analogous to Solomonoff induction.

First we need to define our version of Kolmogorov complexity, which is quite simply

\[
I(s) = \min_{\pi(c) = s} |c|
\]

That is, $I = \min \circ || \circ \pi^{-1}$. We will call I the "index complexity" since it is the size of the smallest indexing circuit. Note that I is the minimum circuit size in the sense of the well known minimum circuit size problem (MCSP). We can extend I to $\mathcal{P}(\mathbb{B}^n)$ as
\[
I(x) = \min_{s \in x} I(s)
\]
Note that this definition can be read as stating that I(x) is the size of the smallest circuit that computes the bits of x when given their indices as input, with behavior on any out of range indices unspecified. Determining I(x) from x is roughly equivalent to the Partial-MCSP. The definition is natural because I(x) is the complexity of the simplest explanation for the data given by x. 

This definition of complexity is not equivalent to Kolmogorov's. In particular, a short program for a string x cannot necessarily be translated into a small circuit, since the Cook-Levin construction would roll a computation history out to a very large circuit if the runtime is large. So we might expect that sometimes I(x) > K(x). In fact, a short program to output x must take at least |x| steps, so translating it into a circuit and then indexing the output to find $x_i$ always results in a larger circuit than we can find by treating x as a Boolean function and compressing its circuit. Generating the entire string is in fact wasteful from the perspective of index complexity; the natural analogue is really to the shortest program that produces each bit of x given its index, which is potentially much faster than the shortest program to produce x. The minimum circuit size required to compute a string x of length $2^n$, what we call I(x), is related to the index-wise version of the time bounded Kolmogorov complexity, KT(x), as shown in \cite{WorldsCollide} (see theorem 9). Allender does not introduce a term for I(x), and the bounds are not as tight as a constant factor, so it is not clear whether KT(x) and I(x) capture the same idea. Since this would be quite a coincidence, we suspect not. The close connection between Kolmogorov complexity and the MCSP has been observed explicitly \cite{KolmogorovAndMCSP} in the past. Despite extensive work relating the two areas, we are not aware of anyone suggesting circuit size as a basis for an alternative algorithmic information theory.   

\section{Effectiveness of Minimum Index Complexity Prediction}

\subsection{The MIP} \label{MIP}

Our conjectured relationship between $\mu$ and I is that predicting the bits of a string s sampled from $\mu$ based on the smallest circuit computing the bits seen so far usually makes few errors.

Formally, let the Minimum Index complexity Predictor (MIP) produce predictions for the next bit as follows:
\[
\MIP(s) = \argmin_b I(sb)
\]
Generally the MIP produces a set of possible predictions. These are the next bit(s) belonging to strings computed by the smallest circuit computing s. In fact, $I(s*) = \min_b I(sb)$ since if c is some smallest circuit with $\pi(c) \in s*$, and letting $\pi(c) = sb$, I(sb) is no greater than $|c| = I(s*)$. Therefore we can think of the MIP as constructing the smallest circuit computing the bits we have seen and using it to predict the next bit. For now we will ignore the issue of what prediction the MIP should make when the RHS has cardinality 2 (when two minimal circuits give different predictions). One might imagine that the MIP flips a coin.

Then our goal is to prove that for strings x sampled from $\mu$, usually $x_k \in \MIP(x_{<k})$.

\subsection{MIP Performance Bounds} \label{MIP Proof}

Let $C_g = ||^{-1}(g)$ be the set of circuits of size g. Note that
\[
1 = \lambda(\Lambda) = \lambda(\cup_{g \in \mathcal{G}} C_g) = \sum_{g \in \mathcal{G}} |C_g| f(g)
\]
Therefore, $\lambda$ implicitly defines a measure $\nu : \mathcal{G} \rightarrow \mathbb{R}$ by $\nu(g) = |C_g|f(g)$. For a family of measures $\lambda_n$ for different input sizes $\lg n$, the behavior of $\nu$ as $n \rightarrow \infty$ will later tell us about the properties of the corresponding $\mu_n$. We can think of $\nu(g)$ as the chance of sampling a size g circuit.  

Though $C_g$ are the "level sets" of $\Lambda$, they are not directly connected to I. To understand the MIP, we must introduce
\[
X_g = I^{-1}(g)
\]
the set of strings of index complexity g. Now, all strings have some index complexity, since any string is computed by a sufficiently large circuit (for instance with the disjunctive normal form), which means $\{X_g\}_{g \in \mathcal{G}}$ form a partition of $\mathbb{B}^n$. And of course it is clear that $\{C_g\}_{g\in \mathcal{G}}$ form a partition of $\Lambda$. 

The connection between these partitioned sets is of course circuit computation, $\pi$, a surjective function that is very far from injective. 

Directly from the definitions,
\[
\pi^{-1}(X_g) \subseteq \cup_{k=g}^\infty C_k
\]
that is, if I(x) = g it is computed only by circuits of size at least g. Therefore $\pi^{-1}$ maps a string x to a "cone" rooted at $C_{I(x)}$ and cutting through $C_k$ for all larger k. 

It follows that
\[
\pi^{-1}(\cup_{k=g}^\infty X_k) \subseteq \cup_{k=g}^\infty C_k
\]
In the other direction, 
\[
C_g \subseteq \pi^{-1}(\cup_{k=0}^g X_k)
\]
because a size g circuit can only compute strings of index complexity at most g. But this means that
\[
\cup_{k=0}^g C_k \subseteq \pi^{-1}(\cup_{k=0}^g X_k)
\]

Now we can see that the probability a random string x with distribution $\mu$, ($x \sim \mu$), has index complexity at most N is
\[
P_{x \sim \mu}(I(x) \leq N) = \mu(\cup_{k=0}^N X_k) = \lambda (\pi^{-1}(\cup_{k=0}^g X_g)) \geq \lambda(\cup_{k=0}^g C_k) = \sum_{k=0}^N |C_k|f(k) = \sum_{k=0}^N \nu(k)
\]

which converges to 1 as $N \rightarrow \infty$. 

\begin{theorem} \label{ErrorNumberBound}
    The $\mu$-probability that $|\{ k | x_{k} \notin \MIP(x_{<k}) \}| \leq N$ is at least $\sum_{k=0}^N \nu(k)$.
\end{theorem}

We will call the size of the set in the theorem the number of errors of the MIP predictor on x. It is important to remember that the MIP may be "uncertain" even when it is not in error. 

Proof: For $x \sim \mu$, we have shown that $P(I(x) \leq N) \geq \sum_{k=0}^N \nu(k)$. We will consider the maximum number of errors of the MIP predictor for x in this set. Let $N_k = I(x_{\leq k})$ be size of the smallest circuit computing the length k prefix of x. If $x_k \notin \MIP(x_{<k})$, then because $I(x_{<k}b) \geq I(x_{<k})$, $N_{k-1} = I(x_{<k}) < I(x_{\leq k}) = N_k$. In other words, where the MIP is wrong, $N_k$ increases. In fact, because the index complexity is integral valued, $N_k \geq N_{k-1} + 1$. But $N_0 = I(*) = 0$ and $N_n = I(x) \leq N$, so the number of errors is at most N. 

In fact, there is a corresponding negative result. We can always "hardcode" a particular bit position with no more than $2 \lg n$ gates by checking that each bit of the index is set to the correct value, so $I(xb) \leq I(x) + 2 \lg n$. But this means that $N_k < N_{k-1} + 2\lg n$. Since $N_0$ must increase to $N_n = I(x)$ in steps of size at most $2 \lg n$, the MIP makes at least $\frac{I(x)}{2\lg n}$ prediction errors on the string x. 

Now, we would like to argue that for large n the probability of few prediction errors goes to 1. For this to be meaningful, we must actually consider a sequence of circuit priors $\mu_n$ on input size n (circuits are a nonuniform model of computation, so we cannot use one fixed circuit prior for all string lengths). To show that the probability of few errors goes to 1 using the result in theorem $\ref{ErrorNumberBound}$, we must allow N to approach infinity. But from the proof, the maximum number of errors for x with $I(x) \leq N$ is N. The fraction of errors among all n predictions is then $\frac{N}{n}$, and we desire for this to approach 0. Then, intuitively, we cannot allow the probability mass of $\mu_n$ to "escape to the tail" of infinite circuit size - it must be possible to increase N sublinearly in n, and still maintain the requirement that $\sum_{k=0}^N \nu_n(k) \rightarrow 1$, where $\nu_n$ corresponds to $\mu_n$. If this is true, we say that $\mu_n$ is $\nu$-stable. 

Formally, a family of circuit priors $\mu_n$ induced by $\nu_n$ is $\nu$-stable if there exists a sequence $g_n \in o(n)$ such that $\sum_{k=0}^{g_n} \nu_n \rightarrow 1$.  

Then, as intended by the definition of $\nu$-stability, we have the following result:

\begin{theorem} \label{ErrorFractionBound}
    For a $\nu$-stable family of circuit priors $\mu_n$, and any $\epsilon > 0$, the probability that the MIP makes no more than $\epsilon n$ errors approaches 1 as $n \rightarrow \infty$. 
\end{theorem}

Proof: Assume $\mu_n$ is $\nu$-stable, and consider any $\epsilon > 0$. Then, allowing $g_n$ to satisfy the criteria of $\nu$-stability, we have directly from theorem \ref{ErrorNumberBound} that the MIP makes no more than $g_n$ errors on a sequence of length n with probability at least $\sum_{k=0}^{g_n} \nu_n \rightarrow 1$. But $g_n \in o(n)$, so $\frac{g_n}{n} \rightarrow 0$. Therefore $\exists n_0$ s.t. $\forall n > n_0$, $\frac{g_n}{n} < \epsilon$. But this implies $g_n < \epsilon n$, providing the desired error bound.  

The question remains of how frequently the MIP is uncertain. A weak upper bound is fairly easy to obtain: any time the MIP is uncertain between (w.l.o.g.) the true continuation $x_{<k}0$ and $x_{<k}1$, there must be some string $z_k = x_{<k}1y$ with $I(z_k) \leq I(x)$. Therefore, if $x \in L_g = \cup_{k=0}^g X_k$, certainly $z_k \in \cup_{k=0}^g X_k$. Let $U \subseteq \{0, .., n-1\}$ be this set of indices where the MIP is uncertain. Then we see that $\forall i \in U$, $\exists z_i$ with $z_i \in L_g$. In particular, $z_i$ is identical to x up to index i, where it differs, so $z_i \neq z_j$ for $i \neq j$. This means $|U| = |\{z_i\}_{i \in U}| \leq |L_g|$. The MIP cannot be uncertain more than $|L_g| =|\cup_{k=0}^g X_k|$ times.

Now $|L_g|$ is polylogarithmic in n, since the number of inputs is $\lg n$, and the number of circuits with g gates is upper bounded by $(g + \lg n)\prod_{k=0}^{g-1} (k + \lg n)^2$ \ref{CircuitSize}. But this grows very quickly in g, so $\nu$-stability may not be enough to ensure that the probability of few uncertain indices goes to 1. In the restricted case when $\nu_n$ is constant we can say this - we will call this the requirement that $\mu_n$ is $\nu$-constant, so that

\begin{theorem} \label{CorrectnessFractionBound}
    If $\mu_n$ is $\nu$-constant, then for any $\epsilon > 0$, the probability that the MIP is correct at least $(1-\epsilon)n$ times approaches 1 as $n \rightarrow \infty$.
\end{theorem}

Proof: If $\mu_n$ is $\nu$-constant it is certainly $\nu$-stable, so theorem \ref{ErrorFractionBound} applies. This gives us a bound of $\frac{\epsilon}{2}n$, errors. Allow $g_n$ to approach $\infty$ sufficiently slowly that $|L_{g_n}|$ grows sublinearly. Then for sufficiently large n we obtain less than $\frac{\epsilon}{2}n$ uncertain indices, and theorem \ref{CorrectnessFractionBound} follows.

The requirement of $\nu$-constancy is actually quite natural. Re-arranging the definition of $\nu(g)$, $f(g) = \frac{\nu(g)}{|C_g|}$. Scaling f inversely with $|C_g|$ gives a $\nu$-constant family $\mu_n$. This corresponds to sampling a circuit size g with fixed probability, then sampling a circuit uniformly from $C_g$. When constructing the circuit prior, the authors had in mind $\mu(g) = 2^{-(g+1)}$. 

The requirement of $\nu$-constancy may be weakened to $\nu$-stability by taking advantage of the observation that clearly not all strings in $L_g$ can be obtained through the construction of z (assuming only one string z is constructed at each index). For later indices k presumably most strings in $L_g$ are not prefixed by $x_{<k}$.

\section{Conclusion}

We have constructed an algorithmic information theory based on circuits, which is simply a Bayesian mixture over circuits. Because circuits are the most important uniform model of computation in computer science this is a natural prior for reasoning about fixed length objects, such as Boolean functions. We have shown that there are many degrees of freedom in specifying a reasonable circuit prior, which are encapsulated by $\nu$, the choice of prior on the circuit size. This is somewhat similar to the choice of UTM for Solomonoff induction, though the index complexity itself has essentially no degrees of freedom in its specification (unlike the Kolmogorov complexity). We have not shown any invariance under choice of $\nu$, but we have verified properties of the minimum index complexity predictor reminiscent of the MDL principle. Our theory does not extend easily to the prediction of infinite sequences, or sequences of unknown length; in fact our preliminary investigations of possible limiting behaviors have only led to trivial measures, and have been left to future work. However it should be noted that humans usually need to make predictions only about finite sequences (since our lifespans are presumably limited).  

We have noted some advantages of the index complexity over Kolmogorov complexity as a basis for algorithmic information theory. The index complexity is at least computable, and the same can be said for the MIP (both by enumeration of circuits). The circuit prior is clearly lower semicomputable (enumerating circuits up to a maximum circuit size), and since it is a measure it is also estimable (the term "computable" is elsewhere used to mean estimable) \cite{livitanyi}. This cannot be said for the universal a priori distribution (Lemma 4.3.2, \cite{livitanyi}). 

The circuit complexity also has advantages as a basis for studying applications of AIT to machine learning. The highly successful paradigm of deep learning arguably bares more resemblance to non-uniform models of computation like circuits than uniform models like Turing machines, at least for discriminative methods. Indeed the authors believe that deep learning is a search over the space of circuits, and it is easier to search this space by gradient descent than the space of programs or TMs. This motivates an experimental study of whether deep learning (by e.g. backpropogation) samples Boolean functions (approximately) according to the circuit prior. It would also be interesting to investigate whether the relationship between neural networks and Bayesian neural networks \cite{BayesianNN} is similar to the relationship between the MIP and the circuit prior. 

Finally, we suggest that perhaps circuits also have philosophical advantages over TMs as a model of reality. The laws of physics in our universe are highly local, and it is reasonable to imagine that a distributed, parallel model of computation is appropriate to describe them. One alternative model that deserves investigation for this purpose is the cellular automata \cite{automata}. 

\section{Future Work} \label{FutureWork}

In the future, we would like to improve the bound in theorem \ref{CorrectnessFractionBound}, which we suspect can be done by exhibiting, for a given string x, some small circuits which are not obtained by the process described in the proof. Interestingly few of the results in this paper actually depend on the nature of circuit computation, which was an intentional choice in view of the difficulty of finding circuit lower bounds, but suggests that results may be sharpened by taking advantage of circuit upper bounds. 

It may be possible to motivate a particular choice of $\nu$ based on desirable features (such as a regular rate of decay), and to investigate the dependence of $\mu$ on $\nu$ as bits are observed. It is not clear how strongly the predictions of $\mu$ depend on the choice of $\nu$; one might hope that observing enough data soon makes the choice of prior irrelevant.

We would also like to find a natural family $\mu_n$ limiting to a continuous measure $\mu_\infty : \mathcal{P}(\mathbb{Q}) \rightarrow \mathbb{R}$. This would require more measure theory than we used in this paper; in particular we would need a measure $\lambda_\infty$ on the $\sigma$-algebra generated by $\pi$ (appropriately extending $\pi$ to infinite input circuits), and also factoring through ||, so that $\lambda_\infty = f \circ ||$.


\printbibliography

\section{Appendix}

\subsection{Bounds on $|C_g|$} \label{CircuitSize}

We can obtain an upper bound on the number of g gate, $\lg n$ input circuits with one output by giving a construction for all such circuits. For each gate from 1 to g, add it to the circuit and select the two existing gates or inputs that should be connected to it. These need not be unique. Clearly there are at most $g-1 + \lg n$ choices for each of the first and second connections, for a total of $(g-1+\lg n)^2$ choices. In fact this overcounts by roughly a factor of two since the order of connections does not matter, but we only seek an upper bound. Finally, select any gate or input gate as the output, from among $g + \lg n$ choices. The upper bound is 

\[
(g+ \lg n) \prod_{k=1}^g (k-1 + \lg n)^2 = (g + \lg n ) \prod_{k=0}^{g-1} (k + \lg n)^2 
\]

We can also obtain a lower bound by a set of choice leading to unique circuits. In particular, we will add gates so that the depth of each successive gate added is always one greater than the last. It's clear that for all but the first gate added (which has two choices), there is one freely chosen connection and one fixed connection (to the previous gate added), and that there are no nontrivial automorphisms of circuits constructed this way (say, as graphs, with inputs fixed), since any gate would have to map to a gate at the same depth, of which there is only one.

This argument gives a very loose lower bound of
\[
(g + \lg n) \prod_{k=1}^g (k-1 + \lg n) = (g + \lg n) \prod_{k=0}^{g-1} (k + \lg n) = \prod_{k=0}^g (k + \lg n) = \frac{(g + \lg n)!}{(\lg n)!}
\]


\subsection{Glossary of Terms} \label{Glossary}

A := $\{0,1,*\}$

$\mathbb{B}$ := $\{0,1\}$

$C_g$ := $||^{-1}(g)$, the set of size g circuits.

$\mathcal{G}$ := an alias for $\mathbb{N}$, used to remind the reader that we have in mind the number of gates in a circuit.  

$I : \mathcal{P}(\mathbb{B}^n) \rightarrow \mathbb{N} $ := the index complexity, an extension of circuit complexity. Sometimes $\mathcal{G}$ is used as an alias for $\mathbb{N}$.  

$\Lambda$ := The set of all circuits on some fixed input size.

$\lambda : \mathcal{P}(\Lambda) \rightarrow [0,1]$ := a measure on the set of circuits. For our purposes, $\lambda$ depends only on the circuit size, so that (viewed as a function on individual circuits) $\lambda = f \circ ||$.  The choice of $\lambda$ determines $\mu$. 

MIP := the minimum index complexity predictor.

$\mu : \mathcal{P}(\mathbb{B}^n) \rightarrow [0,1]$ := the circuit prior.

$\nu : \mathcal{\mathbb{N}} \rightarrow [0,1]$ := the induced measure on the natural numbers, interpreted as circuit sizes, associated with a given circuit prior $\mu$. Sometimes $\mathcal{G}$ is used as an alias for $\mathbb{N}$.

$\pi$ : $\Lambda \rightarrow \mathbb{B}^n$ := circuit computation.

$X_g$ := $I^{-1}(g)$, the set of index complexity g strings (of some length n).

\end{document}